# Tagging accurately – Don't guess if you know

**Pasi Tapanainen**
Rank Xerox Research Centre
Grenoble Laboratory
6, chemin de Maupertuis
38240 Meylan, France
Pasi.Tapanainen@xerox.fr

**Atro Voutilainen**
Research Unit for Computational Linguistics
University of Helsinki
P.O. Box 4
00014 University of Helsinki, Finland
Atro.Voutilainen@helsinki.fi

## Abstract

We discuss combining knowledge-based (or rule-based) and statistical part-of-speech taggers. We use two mature taggers, ENGCG and Xerox Tagger, to independently tag the same text and combine the results to produce a fully disambiguated text. In a 27000 word test sample taken from a previously unseen corpus we achieve 98.5 % accuracy. This paper presents the data in detail. We describe the problems we encountered in the course of combining the two taggers and discuss the problem of evaluating taggers.

## 1 Introduction

This paper combines knowledge-based and statistical methods for part-of-speech disambiguation, taking advantage of the best features of both approaches. The resulting output is fully and accurately disambiguated.

We demonstrate a system that accurately resolves most part-of-speech ambiguities by means of syntactic rules and employs a stochastic tagger to eliminate the remaining ambiguity. The overall results are clearly superior to the reported results for state-of-the-art stochastic systems.

The input to our part-of-speech disambiguator consists of lexically analysed sentences. Many words have more than one analysis. The task of the disambiguator is to select the contextually appropriate alternative by discarding the improper ones.

Some of the inappropriate alternatives can be discarded reliably by linguistic rules. For example, we can safely exclude a finite-verb reading if the previous word is an unambiguous determiner. The application of such rules does not always result in a fully disambiguated output (e.g. adjective–noun ambiguities may be left pending) but the amount of ambiguity is reduced with next to no errors. Using a large collection of linguistic rules, a lot of ambiguity can be resolved, though some cases remain unresolved.

The rule system may also exploit the fact that certain linguistically possible configurations have such a low frequency in certain types of text that they can be ignored. A rule that assumes that a preposition is followed by a noun phrase may be a useful heuristic rule in a practical system, considering that dangling prepositions occur relatively infrequently. Such heuristic rules can be applied to resolve some of the ambiguities that survive the more reliable grammar rules.

A stochastic disambiguator selects the most likely tag for a word by consulting the neighbouring tags or words, typically in a two or three word window. Because of the limited size of the window, the choices made by a stochastic disambiguator are often quite naive from the linguistic point of view. For instance, the correct resolution of a preposition vs. subordinating conjunction ambiguity in a small window is often impossible because both morphological categories can have identical local contexts (for instance, both can be followed by a noun phrase). Some of the errors made by a stochastic system can be avoided in a knowledge-based system because the rules can refer to words and tags in the scope of the entire sentence.

We use both types of disambiguators. The knowledge-based disambiguator does not resolve all ambiguities but the choices it makes are nearly always correct. The statistical disambiguator resolves all ambiguities but its decisions are not very reliable. We combine these two disambiguators; here this means that the text is analysed with both systems. Whenever there is a conflict between the systems, we trust the analysis proposed by the knowledge-based system. Whenever the knowledge-based system leaves an ambiguity unresolved, we select that alternative which is closest to the selection made by the statistical system.

The two systems we use are ENGCG (Karlsson et al., 1994) and the Xerox Tagger (Cutting et al., 1992). We discuss problems caused by the fact that these taggers use different tag sets, and present the results obtained by applying the combined taggers to a previously unseen sample of text.

## 2 The taggers in outline

### 2.1 English Constraint Grammar Parser

The English Constraint Grammar Parser, ENGCG (Voutilainen et al., 1992; Karlsson et al., 1994), is based on Constraint Grammar, a parsing framework proposed by Fred Karlsson (1990). It was developed 1989–1993 at the Research Unit for Computational Linguistics, University of Helsinki, by Atro Voutilainen, Juha Heikkilä and Arto Anttila; later on, Timo Järvinen has extended the syntactic description, and Pasi Tapanainen has made a new fast implementation of the CG parsing program. ENGCG is primarily designed for the analysis of standard written English of the British and American varieties. In the development and testing of the system, over 100 million words of running text have been used.

The ENGTWOL lexicon is based on the two-level model (Koskenniemi, 1983). The lexicon contains over 80,000 lexical entries, each of which represents all inflected and central derived forms of the lexemes. The lexicon also employs a collection of tags for part of speech, inflection, derivation and even syntactic category (e.g. verb classification).

Usually less than 5 % of all word-form tokens in running text are not recognised by the morphological analyser. Therefore the system employs a rule-based heuristic module that provides all unknown words with one or more readings. About 99.5 % of words not recognised by the ENGTWOL analyser itself get a correct analysis from the heuristic module. The module contains a list of prefixes and suffixes, and possible analyses for matching words. For instance, words beginning with *un...* and ending in *...al* are marked as adjectives.

The grammar for morphological disambiguation (Voutilainen, 1994) is based on 23 linguistic generalisations about the form and function of essentially syntactic constructions, e.g. the form of the noun phrase, prepositional phrase, and finite verb chain. These generalisations are expressed as 1,100 highly reliable 'grammar-based' and some 200 less reliable add-on 'heuristic' constraints, usually in a partial and negative fashion. Using the 1,100 best constraints results in a somewhat ambiguous output. Usually there are about 1.04–1.07 morphological analyses per word. Usually at least 997 words out of every thousand retain the contextually appropriate morphological reading, i.e. the recall usually is at least 99.7 %. If the heuristic constraints are also used, the ambiguity rate falls to 1.02–1.04 readings per word, with an overall recall of about 99.5 %. This accuracy compares very favourably with results reported in (de Marcken, 1990; Weischedel et al., 1993; Kempe, 1994) – for instance, to reach the recall of 99.3 %, the system by (Weischedel et al., 1993) has to leave as many as three readings per word in its output.

### 2.2 Xerox Tagger

The Xerox Tagger[1], XT, (Cutting et al., 1992) is a statistical tagger made by Doug Cutting, Julian Kupiec, Jan Pedersen and Penelope Sibun in Xerox PARC. It was trained on the untagged Brown Corpus (Francis and Kučera, 1982).

The lexicon is a word-list of 50,000 words with alternative tags. Unknown words are analysed according to their suffixes. The lexicon and suffix tables are implemented as tries. For instance, for the word *live* there are the following alternative analyses: *JJ* (adjective) and *VB* (uninflected verb). Unknown words not recognised by suffix tables get all tags from a specific set (called open-class).

The tagger itself is based on the Hidden Markov Model (Baum, 1972) and word equivalence classes (Kupiec, 1989). Although the tagger is trained with the untagged Brown corpus, there are several ways to 'force' it to learn.

- The symbol biases represent a kind of lexical probabilities for given word equivalence classes.
- The transition biases can be used for saying that it is likely or unlikely that a tag is followed by some specific tag. The biases serve as default values for the Hidden Markov Model before the training.
- Some rare readings may be removed from the lexicon to prevent the tagger from selecting them.
- There are some training parameters, like the number of iterations (how many times the same block of text is used in training) and the size of the block of the text used for training.
- The choice of the training corpus affects the result.

The tagger is reported (Cutting et al., 1992) to have a better than 96 % accuracy in the analysis of parts of the Brown Corpus. The accuracy is similar to other probabilistic taggers.

## 3 Grammatical representations of the taggers

A major difference between a knowledge-based and a probabilistic tagger is that the knowledge-based tagger needs as much information as possible while the probabilistic tagger requires some compact set of tags that does not make too many distinctions between similar words. The difference can be seen by comparing the Brown Corpus tag set (used by XT) with the ENGCG tag set.

The ENGTWOL morphological analyser employs 139 tags. Each word usually receives several tags (see Figure 1). There are also 'auxiliary' tags for derivational and syntactic information that do not increase morphological ambiguity but serve as additional information for rules. If these auxiliary tags are ignored, the morphological analyser produces about 180 different tag combinations.

---

[1] We use version 1.

| | ENGCG | XT |
|---|---|---|
| has | V PRES SG3 VFIN | hvz |
| have | V PRES -SG3 VFIN<br>V INF<br>V IMP VFIN<br>V SUBJUNCTIVE VFIN | hv |
| was | V PAST SG1,3 VFIN | bedz |
| do | V PRES -SG3 VFIN<br>V INF<br>V IMP VFIN<br>V SUBJUNCTIVE VFIN | do |
| done | PCP2 | vbn |
| cook | V PRES -SG3 VFIN<br>V INF<br>V IMP VFIN<br>V SUBJUNCTIVE VFIN<br>N NOM SG | vb<br><br><br><br>nn |
| cool | V PRES -SG3 VFIN<br>V INF<br>V IMP VFIN<br>V SUBJUNCTIVE VFIN<br>A ABS | vb<br><br><br><br>jj<br>nn<br>rb |
| cooled | PCP2<br>V PAST VFIN | vbn<br>vbd |
| cooling | PCP1 | vbg<br>nn |

Figure 1: Some morphological ambiguities for verbs.

| Brown tag | Two most probable ENGCG tags (%) |
|---|---|
| CS | CS (70 %)<br>PREP (28 %) |
| DT | DET DEM SG (48 %)<br>PRON DEM SG (27 %) |
| DTI | DET SG/PL (68 %)<br>PRON SG/PL (28 %) |
| IN | PREP (99 %)<br>ADV (0.5 %) |
| JJ | A ABS (93 %)<br>N NOM SG (3 %) |
| NN | N NOM SG (88 %)<br>N NOM SG/PL (7 %) |
| NP | N NOM SG (80 %)<br>N NOM PL (7 %) |
| VB | V INF (84 %)<br>V PRES -SG3 VFIN (12 %) |
| * | NEG-PART (100 %) |

Figure 2: Some mappings from the Brown Corpus to the ENGCG tagset.

The XT lexicon contains 94 tags for words; 15 of them are assigned unambiguously to only one word. There are 32 verb tags: 8 tags for *have*, 13 for *be*, 6 for *do* and 5 tags for other verbs. ENGCG does not make a distinction in the tagset between words *have*, *be*, *do* and the other verbs. To see the difference with ENGCG, see Figure 1.

The ENGCG description differs from the Brown Corpus tag set in the following respects. ENGCG is more distinctive in that a part of speech distinction is spelled out (see Figure 2) in the description of

- determiner–pronoun homographs,
- preposition–conjunction homographs,
- determiner–adverb–pronoun homographs, and
- uninflected verb forms (see Figure 1), which are represented as ambiguous due to the subjunctive, imperative, infinitive and present tense readings.

On the other hand, ENGCG does not spell out part-of-speech ambiguity in the description of

- *-ing* and nonfinite *-ed* forms,
- noun–adjective homographs when the core meanings of the adjective and noun readings are similar,
- ambiguities due to proper nouns, common nouns and abbreviations.

## 4 Combining the taggers

In our approach we apply ENGCG and XT independently. Combining the taggers means aligning the outputs of the taggers and transforming the result of one tagger to that of the other.

Aligning the output is straightforward: we only need to match the word forms in the output of the taggers. Some minor problems occur when tokenisation is done differently. For instance, XT handles words like *aren't* as a single token, when ENGCG divides it to two tokens, *are* and *not*. Also ENGCG recognises some multiple word phrases like *in spite of* as one token, while XT handles it as three tokens.

We do not need to map both Brown tags to ENGCG and vice versa. It is enough to transform ENGCG tags to Brown tags and select the tag that XT has produced, or transform the tag of XT into ENGCG tags. We do the latter because the ENGCG tags contain more information. This is likely to be desirable in the design of potential applications.

There are a couple of problems in mapping:

- Difference in distinctiveness. Sometimes ENGTWOL makes a distinction not made by the Brown tagset; sometimes the Brown tagset makes a distinction not made by ENGTWOL (see Figure 2).
- Sometimes tags are used in a different way. A

case in point is the word *as*. In a sample of 76 instances of *as* from the tagged Brown corpus, 73 are analysed as *CS*; two as *QL* and one as *IN*, while in the ENGCG description the same instances of *as* were analysed 15 times as *CS*, four times as *ADV*, and 57 times as *PREP*.

In ENGCG, the tag *CS* represents subordinating conjunctions. In the following sentences the correct analysis for word *as* in ENGCG is *PREP*, not *CS*, which the Brown corpus suggests.

> The city purchasing department, the jury said, is lacking in experienced clerical personnel as(CS) a result of city personnel policies. — The petition listed the mayor's occupation as(CS) attorney and his age as(CS) 71. It listed his wife's age as(CS) 74 and place of birth as(CS) Opelika, Ala.

The sentences are the three *first* sentences where word *as* appears in Brown corpus. In the Brown Corpus *as* appears over 7000 times and it is the fourteenth most common word. Because XT is trained according to the Brown Corpus, this is likely to cause problems.

XT is applied independently to the text, and the tagger's prediction is consulted in the analysis of those words where ENGCG is unable to make a unique prediction. The system selects the ENGCG morphological reading that most closely corresponds to the tag proposed by XT.

The mapping scheme is the following. For each Brown Corpus tag, there is a decision list for possible ENGCG tags, the most probable one first. We have computed the decision list from the part of Brown Corpus that is also manually tagged according to the ENGCG grammatical representation. The mapping can be used in two different ways.

- *Careful mode*: An ambiguous reading in the output of ENGCG may be removed only when it is not in the decision list. In practise this leaves quite much ambiguity.

- *Unambiguous mode*: Select the reading in the output of ENGCG that comes first in the decision list[2].

## 5 Performance test

### 5.1 Test data

The system was tested against 26,711 words of newspaper text from *The Wall Street Journal*, *The Economist* and *Today*, all taken from the 200-million word *Bank of English* corpus by the COBUILD team at the University of Birmingham, England (see also (Järvinen, 1994)). None of these texts have been used in the development of the system or the description, i.e. no training effects are to be expected.

---

[2] In some cases a word may still remain ambiguous.

### 5.2 Creation of benchmark corpus

Before the test, a benchmark version of the test corpus was created. The texts were first analysed using the preprocessor, the morphological analyser, and the module for morphological heuristics. This ambiguous data was then manually disambiguated by judges, each having a thorough understanding of the ENGCG grammatical representation. The corpus was independently disambiguated by two judges. In the instructions to the experts, special emphasis was given to the quality of the work (there was no time pressure). The two disambiguated versions of the corpus were compared using the Unix sdiff program. At this stage, slightly above 99 % of all analyses agreed. The differences were jointly examined by the judges to see whether they were caused by inattention or by a genuine difference of opinion that could not be resolved by consulting the documentation that outlines the principles adopted for this grammatical representation (for the most part documented in (Karlsson et al., 1994)). It turned out that almost all of these differences were due to inattention. Only in the analysis of a few words it was agreed that a multiple choice was appropriate because of different meaning-level interpretations of the utterance (these were actually headings where some of the grammatical information was omitted). Overall, these results agree with our previous experiences (Karlsson et al., 1994): if the analysis is done by experts in the adopted grammatical representation, with emphasis on the quality of the work, a consensus of virtually 100 % is possible, at least at the level of morphological analysis (for a less optimistic view, see (Church, 1992)).

### 5.3 Morphological analysis

The preprocessed text was submitted to the ENG-TWOL morphological analyser, which assigns to 25,831 words of the total 26,711 (96.7 %) at least one morphological analysis. The remaining 880 word-form tokens were analysed with the rule-based heuristic module. After the combined effect of these modules, there were 47,269 morphological analyses, i.e. 1.77 morphological analyses for each word on an average. At this stage, 23 words missed a contextually appropriate analysis, i.e. the error rate of the system after morphological analysis was about 0.1 %.

### 5.4 Morphological disambiguation

The morphologically analysed text was submitted to five disambiguators (see Figure 3). The first one, D1, is the grammar-based ENGCG disambiguator. In the next step (D2) we have used also heuristic ENGCG constraints. The probabilistic information is used in D3, where the ambiguities of D2 are resolved by XT. We also tested the usefulness of the heuristic component of ENGCG by omitting it in D4. The last test, D5, is XT alone, i.e. only probabilistic techniques are used here for resolving ENGTWOL ambiguities.

The ENGCG disambiguator performed somewhat less well than usually. With heuristic constraints, the error rate was as high as 0.63 %, with 1.04 morphological readings per word on an average. However, most (57 %) of the total errors were made *after* ENGCG analysis (i.e. in the analysis of no more than 3.6 % of all words). In a way, this is not very surprising because ENGCG is supposed to tackle all the 'easy' cases and leave the structurally hardest cases pending. But it is quite revealing that as much as three fourths of the probabilistic tagger's errors occur in the analysis of the structurally 'easy' cases; obviously, many of the probabilistic system's decisions are structurally somewhat naive. Overall, the hybrid (D3$_\beta$) reached an accuracy of about 98.5 % – significantly better than the 95–97 % accuracy which state-of-the-art probabilistic taggers reach alone.

The hybrid D3$_\alpha$ is like hybrid D3$_\beta$, but we have used careful mapping. There some problematic ambiguity (see Figure 2) is left pending. For instance, ambiguities between preposition and infinitive marker (word *to*), or between subordinator and preposition (word *as*), are resolved as far as ENGCG disambiguates them, the prediction of XT is not consulted. Also, when XT proposes tags like *JJ* (adjective), *AP* (post-determiner) or *VB* (verb base-form) very little further disambiguation is done. This hybrid does not contain any mapping errors, and on the other hand, not all the XT errors either.

The test without the heuristic component of ENGCG (D4) suggests that ambiguity should be resolved as far as possible with rules. An open question is, how far we can go using only linguistic information (e.g. by writing more heuristic constraints to be applied after the more reliable ones, in this way avoiding many linguistically naive errors).

The last test gives further evidence for the usefulness of a carefully designed linguistic rule component. Without such a rule component, the decrease in accuracy is quite dramatic although a part of the errors come from the mapping between tag sets[3].

## 6   Conclusion

In this paper we have demonstrated how knowledge-based and statistical techniques can be combined to improve the accuracy of a part of speech tagger. Our system reaches a better than 98 % accuracy using a relatively fine-grained grammatical representation.

Some concluding remarks are in order.

---

[3]Even without the mapping errors, the reported 4 % error rate of XT is considerably higher than that of our hybrid.

- Using linguistic information before a statistical module provides a better result than using a statistical module alone.
- ENGCG leaves some 'hard' ambiguities unresolved (about 3–7 % of all words). This amount is characteristic of the ENGCG rule-formalism, tagset and disambiguation grammar. It does not necessarily hold for other knowledge-based systems.
- Only about 20–25 % of errors made by the statistical component occur in the analysis of these 'hard' ambiguities. That means, 75–80 % of the errors made by the statistical tagger were resolved correctly using linguistic rules.
- Certain kinds of ambiguity left pending by ENGCG, e.g. CS vs. PREP, are resolved rather unreliably by XT.
- The overall result is better than other state-of-the-art part-of-speech disambiguators. In our 27000 word test sample from previously unseen corpus, 98.5 % of words received a correct analysis. In other words, the error rate is reduced at least by half.

Although the result is better than provided by any other tagger that produces fully disambiguated output, we believe that the result could still be improved. Some possibilities:

- We could use partly disambiguated text (e.g. the output of parsers D1, D2 or D3$_\alpha$) and disambiguate the result using a knowledge-based syntactic parser (see experiments in (Voutilainen and Tapanainen, 1993)).
- We could leave the text partly disambiguated, and use a syntactic parser that uses both linguistic knowledge and corpus-based heuristics (see (Tapanainen and Järvinen, 1994)).
- Some ambiguities are very difficult to resolve in a small window that statistical taggers currently use (e.g. CS vs. PREP ambiguity when a noun phrase follows). A better way to resolve them would probably be to write (heuristic) rules.
- We could train the statistical tagger on the output of a knowledge-based tagger. That is problematic because generally statistical methods seem to require some compact set of tags, while a knowledge-based system needs more informative tags. The tag set of a knowledge-based system should be reduced down to some subset. That might prevent some mapping errors but there is no quarantee that the statistical tagger would work any better.
- We could try the components in a different order: using statistics before heuristical knowledge etc. However, currently the heuristic component makes less errors than the statistical tagger.

|  | Amb. words | Readings | Readings / word | Errors | Error rate (%) |
|---|---|---|---|---|---|
| D0 (Morphological analysis) | 37.6 % | 47269 | 1.77 | 23 | 0.09 % |
| D1 (D0 + ENGCG) | 6.4 % | 28815 | 1.08 | 94 | 0.35 % |
| D2 (D1 + ENGCG heuristics) | 3.6 % | 27681 | 1.04 | 169 | 0.63 % |
| $D3_\alpha$ (D2 + XT + C-mapping) | 2.2 % | 27358 | 1.02 | 220 | 0.82 % |
| $D3_\beta$ (D2 + XT + mapping) | 0.0 % | 26744 | **1.00** | 391 | **1.46 %** |
| D4 (D1 + XT + mapping) | 0.0 % | 26794 | 1.00 | 597 | 2.24 % |
| D5 (D0 + XT + mapping) | 0.7 % | 26977 | 1.01 | 1703 | 6.38 % |

Figure 3: Performance of the taggers on a 26,711-word corpus.

- We could use a better statistical tagger. But the accuracy of XT is almost the same as the accuracy of any other statistical tagger. What is more, the accuracy of the purely statistical taggers has not been greatly increased since the first of its kind, CLAWS1, (Marshall, 1983) was published over ten years ago.

We believe that the best way to boost the accuracy of a tagger is to employ even more linguistic knowledge. The knowledge should, in addition, contain more syntactic information so that we could refer to real (syntactic) objects of the language, not just a sequence of words or parts of speech. Statistical information should be used only when one does not know how to resolve the remaing ambiguity, and there is a definite need to get fully unambiguous output.

## 7 Acknowledgements

We would like to thank Timo Järvinen, Lauri Karttunen, Jussi Piitulainen and anonymous referees for useful comments on earlier versions of this paper.